\documentclass[a4paper,twoside]{article}

\usepackage{epsfig}
\usepackage{subcaption}
\usepackage{calc}
\usepackage{amssymb}
\usepackage{amstext}
\usepackage{amsmath}
\usepackage{amsthm}
\usepackage{multicol}
\usepackage{pslatex}
\usepackage{booktabs}
\usepackage{array}
\usepackage{graphicx}
\usepackage{url}
\usepackage{tikz}
\usetikzlibrary{arrows.meta,positioning}
\usepackage{apalike}
\usepackage{algorithm2e}  
\usepackage[bottom]{footmisc}
\usepackage{SCITEPRESS}
\usepackage[colorlinks=true,allcolors=blue]{hyperref}

\newif\ifcameraready
\camerareadytrue
\newcommand{\authororcid}[1]{\if\relax\detokenize{#1}\relax\else\orcidAuthor{#1}\fi}

\begin{document}

\title{Would an LLM Match into a Residency? Uncertainty, Reasoning and Image Use of Open Language Models on the Spanish MIR Examination}


\ifcameraready
\author{\authorname{Miguel~Fernandez-de-Retana \authororcid{0009-0002-0883-1303}\sup{\dagger}, Unai~Zulaika \authororcid{0000-0002-7366-9579}, Rubén~Sánchez-Corcuera \authororcid{0000-0002-6361-0901}, \\Aritz~Bilbao-Jayo \authororcid{0000-0001-7743-6652} and~Aitor~Almeida \authororcid{0000-0002-1585-4717}}
\affiliation{Faculty of Engineering, University of Deusto (Bilbao, Spain)}
\email{\sup{\dagger}Corresponding author: m.fernandezderetana@deusto.es}
}
\else
\author{\authorname{Anonymous Submission}
\affiliation{}
\email{}
}
\fi

\keywords{Clinical Reasoning, Large Language Models, Medical Question Answering, Open-Weight Models, Selective Prediction, Uncertainty Quantification, Vision-Language Models.}

\abstract{Accuracy on medical examinations is common evidence offered for the clinical use of language models, yet a health system that intends to run an open model on its own hardware needs to know considerably more: whether the model recognizes when it is likely to be wrong, whether medical fine-tuning improves this ability, whether its correct answers rest on sound clinical reasoning, or whether it actually uses the images it is given. We examine these questions for four open models of 8 to 27 billion parameters on the Spanish MIR residency examination, scoring 1,552 questions from eight editions under the exam's own rules. Answering every question, the three larger models would rank among roughly the first 600 to 3,000 candidates, with 41 to 43 specialties still open to them. Abstention, however, never pays, because even their least confident answers are right more often than the 25\% break-even. Medical fine-tuning reduced overconfidence at the cost of accuracy, without improving how well confidence separates right from wrong answers. Moreover, 13--38\% of correct answers rejected a distractor for a clinically unsound reason that confidence did not reflect, and removing or swapping the provided contextual images left accuracy unchanged.}

\onecolumn \maketitle \normalsize \setcounter{footnote}{0} \vfill

\section{\uppercase{Introduction}}
\label{sec:introduction}

Large language models (LLMs) now pass medical licensing examinations, often by a wide margin \cite{Kung2023,Singhal2023,Nori2023,Kasai2023}, and the Spanish residency examination is no exception \cite{GuillenGrima2023}. Most of this evidence, however, comes from large proprietary models accessed through closed APIs, whose weights are not available and whose prompting, decoding or training data remain outside the user's control. A hospital that cannot send patient data beyond its own infrastructure must thus instead turn to \emph{open models} that can run on its own hardware, which also gives it the ability to inspect, control and audit them \cite{Riedemann2024,Thirunavukarasu2023}. At the scale that fits on a single workstation GPU of reasonable cost and memory, i.e., models of a few tens of billions of parameters, accuracy on medical examinations is far from a solved problem, and the question is whether the performance of these models is good enough to support clinical decision-making or education purposes.

Yet, a headline accuracy figure answers none of the questions that a health system would need to ask before deploying such a model for clinical decision support or medical education \cite{Hager2024}. It says nothing about whether the model knows when it is wrong, whether medical fine-tuning improves that self-knowledge or merely changes accuracy, whether a correct answer is reached through reasoning that a physician would consider clinically sound, or whether the model actually looks at the image when a case comes with one. A model can thus be confidently wrong, or right for the wrong reasons, without either showing in its score.

The Spanish \emph{Médico Interno Residente} (MIR) examination makes it possible to address all four questions under realistic conditions. Because it penalizes wrong answers, a candidate's score depends on when they choose to answer or abstain; because it allocates training locations and specialties by rank, a score translates directly into a specialty and hospital where different candidates would be placed under various degrees of competitiveness; and because physicians publish per-option commentaries on many of its questions, and about one question in eight includes a clinical image, it also lends itself to an analysis of reasoning and image use. We evaluate four open models, namely Qwen3-8B, Qwen3-14B, Gemma-3-27B and MedGemma-27B, the last two of which share a base model and differ only in medical tuning, and address the following research questions:

\begin{description}
\item[RQ1.] Scored like a real candidate, where does each model land in the national ranking?
\item[RQ2.] Does medical fine-tuning improve a model's judgement of its own uncertainty, or only its accuracy/practical performance?
\item[RQ3.] When a model gets the answer right, does it reject the wrong options for reasons that a physician would consider clinically sound?
\item[RQ4.] Do vision language models actually use the clinical images they are given, or do they answer instead from the text alone?
\end{description}

\noindent Our results show that a score good enough to win a competitive residency hides substantial weaknesses. The three larger models would rank among the first few thousand candidates and could choose almost any specialty, yet none of the four can profit from the exam's penalty by abstaining, since even their least confident answers are right more often than the 25\% break-even. Medical tuning makes a model less overconfident but also less accurate, and no better at telling its right answers from wrong ones. Between 13 and 38\% of correct answers dismiss at least one wrong option for a clinically unsound reason, which the model's confidence does not reveal, and withholding or swapping the clinical image leaves accuracy unchanged.

In summary, our contributions are: \textit{(i)} a corpus of 2,595 questions extracted from the last twelve official MIR exams, from 2014 to 2025, with answer keys, physician commentary, clinical plates and the published score-to-rank data of the 2023 and 2024 calls; \textit{(ii)} an evaluation under the exam's own scoring and allocation rules of four open models of 8 to 27 billion parameters, with and without medical tuning; \textit{(iii)} a controlled comparison of the effect of medical tuning on calibration, with an exam sat after the model training cutoff as a contamination control; and \textit{(iv)} evidence on model reasoning and image use in questions with physician commentary and clinical plates.

\section{\uppercase{Related Work}}
\label{sec:related}

\noindent\textbf{LLMs on Medical Examinations.} Since ChatGPT approached the pass mark of the USMLE \cite{Kung2023}, medical examinations have become the standard showcase for clinical LLMs. MedQA \cite{Jin2021} and its successors track accuracy \cite{Singhal2023,Singhal2025,Nori2023}, and national examinations such as those of Japan \cite{Kasai2023} and Spain \cite{GuillenGrima2023} have since been added. The MIR in particular is available as a benchmark through HEAD-QA \cite{Vilares2019,CorreaGuillen2026}, and rheumatologists have rated the reasoning of LLMs on its rheumatology questions \cite{MadridGarcia2023}. These studies, however, report accuracy or at most the raw mark, and we are not aware of any that scores a model under the MIR's penalty and allocation rules, or that separates what a model gains by answering from what it could gain by abstaining.

\medskip
\noindent\textbf{Explanations and Reasoning.} A correct answer does not guarantee correct reasoning. GPT-4V was found to give flawed rationales on over a third of the image challenges it answered correctly \cite{JinQ2024}, and the explanations a model produces do not necessarily reflect the process that actually led to its answer \cite{Turpin2023}. To this end, CasiMedicos provides physician-written explanations of MIR questions that cover the wrong options as well as the right one \cite{Alonso2024,Goenaga2024}, which makes it possible to ask whether a model rejects each distractor for a clinically valid reason. We grade that reasoning against the physician's text with an LLM acting as a judge \cite{Zheng2023}.

\medskip
\noindent\textbf{Calibration and Abstention.} For a model's confidence to support deferral, it must be both \textit{calibrated} \cite{Guo2017,Naeini2015} and able to \textit{separate} right from wrong answers \cite{Hendrycks2017}, a trade-off between coverage and risk formalized by selective prediction \cite{Geifman2017}. LLMs do carry some knowledge of their own correctness \cite{Kadavath2022}, although the confidence they verbalize tends to be overinflated and sycophantic \cite{Xiong2024}. It has recently been argued that benchmarks which do not penalize wrong answers reward guessing, and that penalized scoring would instead reward honest abstention \cite{Kalai2025}. The MIR provides such a benchmark, with a penalty that makes it possible to ask whether a model can identify the questions it would answer worse than chance, and whether medical fine-tuning improves that ability.

\medskip
\noindent\textbf{Images and Contamination.} Multimodal models have been shown to score well on medical visual questions while ignoring or misreading the image \cite{Yan2025,JinQ2024}. Examination results are also exposed to training data contamination \cite{Sainz2023}, for which comparing performance before and after a model's training cutoff is a standard control \cite{Roberts2024}.

\section{\uppercase{The MIR Examination}}
\label{sec:exam}

\subsection{Format and Scoring}

The MIR is a single, national, multiple-choice examination sat once a year by every graduating physician who seeks a specialty training post in Spain. Following the Ministry of Health, we name each exam by the year of its call, even though it is sat early in the following year, so the 2025 exam, for instance, took place on 24 January 2026. Papers contained 225 scored questions up to the 2018 exam, 175 in 2019 and 2020 and 200 from 2021 onwards, in each case with 10 reserve questions that replace any question annulled by the examining commission. Questions had five options up to the 2014 exam and four thereafter. According to the official scoring rules, a \textit{correct} answer scores +3, a \textit{wrong} answer $-$1 and a \textit{blank} 0, so that the total mark $S$ is
\begin{equation}
S = 3c - w,
\label{eq:score}
\end{equation}
where $c$ and $w$ are the numbers of correct and incorrect answers, respectively. Published results divide this mark by three, giving the \emph{net score} $S/3 = c - w/3$, i.e., the number of correct answers net of the penalty for wrong ones. We report net scores throughout.

Since answering a question that one gets right with probability $p$ is worth $3p - (1-p) = 4p - 1$ points, against zero for a blank, it follows that
\begin{equation}
\text{Answer} \iff p > \tfrac{1}{4},
\label{eq:breakeven}
\end{equation}
regardless of the number of options. With four options, a random guess therefore has an expected value of exactly zero, and a blank is better than a guess that is right less than 25\% of the time. The relevant question for a model is thus not whether it abstains often enough, but whether it can identify those questions.

\subsection{From Score to Training Post}

Candidates are ranked by a total mark that weighs the examination at 90\% and the academic record at 10\%, with the resulting rank determining the order in which they choose a specialty and a hospital in the national territory, so that each specialty's \textit{cut-off} is the rank of the candidate who took its last place. Since the Ministry of Health stopped publishing per-candidate result lists after the 2021 exam, the only public information consists of the per-specialty summaries that the trade press compiles from the allocation \cite{ConSalud2024,ConSalud2025}. For each specialty, these report the mean \textit{net score} and mean \textit{rank} of its entrants and, for the 2023 exam, the net score and rank of its last entrant, together with the cut-off and the hospital where the last place went.

\section{\uppercase{Data}}
\label{sec:data}

\noindent\textbf{Questions.} We parsed the official Version~0 booklets of the last twelve MIR examinations, from 2014 to 2025, containing 2,595 questions including 120 reserves (Table~\ref{tab:corpus}). Answer keys were taken from the answer tables published by CasiMedicos and from HEAD-QA v2 \cite{CorreaGuillen2026}, matched on question text. Where both sources key the same question they agree on 730 of 773 cases, and 30 of the 43 disagreements fall in the 2017 exam: the four models' unanimous answers agree with it on only 36\% of questions, against 85--93\% in every other year. We therefore start the analysis of RQ1, RQ2 and RQ4 at the 2018 exam, which leaves only four-option papers. Annulled questions are excluded throughout.

\medskip
\noindent\textbf{Physician Commentary.} CasiMedicos also publishes clinician explanations of questions \cite{Alonso2024,Goenaga2024}, and for some of them, the commentary discusses the wrong options one by one. We use the 123 official questions from the 2015 and 2017--2021 exams whose commentary covers at least two options, since only these can reveal whether a distractor was rejected for the right reason.

\medskip
\noindent\textbf{Clinical Plates.} We extracted the clinical plates from the image annexes of the 2019--2025 exams, the only ones available, obtaining 174 questions with a plate, of which 171 are scorable.

\medskip
\noindent\textbf{Specialty Labels.} CasiMedicos further assigns each commented question to a specialty, with 619 of the 1,552 scored questions from 2018--2025 (40\%) carrying such a label.
\medskip

\noindent\textbf{Rank Data.} We built a \textit{score-to-rank} curve for the 2023 exam (88 points, covering mean and last entrants) and for the 2024 exam (45 points, mean entrants only), together with the cut-offs of their 45 and 44 specialties. The 2025 exam has published cut-offs but no scores yet, and for the 2022 exam, we found cut-offs but no per-specialty scores.

\begin{table}[t]
\caption{Question sets used by each research question.}
\label{tab:corpus}
\centering
\small
\begin{tabular}{@{}lrl@{}}
\toprule
Set & Questions & Use \\
\midrule
Official papers, 2014--2025 & 2,595 & Corpus \\
\quad Scored, 2018--2025 & 1,552 & RQ1, RQ2 \\
\quad With commentary on & 123 & RQ3 \\
\qquad two or more options & & \\
\quad With a scorable image plate & 171 & RQ4 \\
\midrule
Rank-curve points, 2023 / 2024 & 88 / 45 & RQ1 \\
Specialty cut-offs, 2023 / 2024 & 45 / 44 & RQ1 \\
\bottomrule
\end{tabular}
\end{table}

\section{\uppercase{Method}}
\label{sec:method}

\begin{figure*}[t]
\centering
\resizebox{\textwidth}{!}{%
\begin{tikzpicture}[
  font=\scriptsize,
  every node/.append style={execute at begin node={\hyphenpenalty=10000\relax}},
  box/.style={draw=black!55, rounded corners=2pt, align=center, text width=3.1cm, minimum height=0.78cm, inner sep=2.5pt, fill=white},
  pass/.style={box, fill=black!6},
  rq/.style={box, draw=black!80, fill=black!14, font=\scriptsize\bfseries},
  arr/.style={-{Latex[length=1.5mm]}, black!65, semithick}]
\node[font=\scriptsize\bfseries] at (0,1.55) {Data};
\node[font=\scriptsize\bfseries] at (4.1,1.55) {Model Runs};
\node[font=\scriptsize\bfseries] at (8.2,1.55) {Scoring};
\node[font=\scriptsize\bfseries] at (12.3,1.55) {Research Questions};
\node[box] (rc) at (0,0.62) {\textbf{Score-to-Rank Curves and}\\\textbf{Specialty Cut-Offs}\\(2023--2024)};
\node[box] (dA) at (0,-0.5) {\textbf{Official Papers} (2018--2025)\\\textit{1,552 Scored Questions}};
\node[pass, minimum height=1.8cm] (pA) at (4.1,-0.5) {\textbf{Pass A: Label Mode}\\Four Models, Full Papers\\Option Probabilities From One Forward Pass};
\node[box] (s1) at (8.2,0) {\textbf{Scoring Under Exam Answering Policies}};
\node[box] (s2) at (8.2,-1.0) {\textbf{ECE, Brier and AUROC}\\Before and After Release};
\node[rq] (q1) at (12.3,0) {RQ1: Net Score\\and Placement};
\node[rq] (q2) at (12.3,-1.0) {RQ2: Medical Tuning\\and Uncertainty};
\node[box] (dB) at (0,-2.15) {\textbf{Physician Commentary on Two or More Options}\\\textit{123 Questions}};
\node[pass] (pB) at (4.1,-2.15) {\textbf{Pass B: Generate Mode}\\Verdict and Reason};
\node[box] (s3) at (8.2,-2.15) {\textbf{Llama-3.3-70B Judge Against Physician Text} Same, Valid or Wrong};
\node[rq] (q3) at (12.3,-2.15) {RQ3: Distractor\\Reasoning};
\node[box] (dC) at (0,-3.3) {\textbf{Clinical Plates}\\\textit{171 Scorable Questions}};
\node[pass] (pC) at (4.1,-3.3) {\textbf{Pass C: Label Mode}\\Two Vision Models\\Own, No or Another Plate};
\node[box] (s4) at (8.2,-3.3) {\textbf{Paired Accuracy per Condition}};
\node[rq] (q4) at (12.3,-3.3) {RQ4: Image Use};
\draw[arr] (dA) -- (pA);
\draw[arr] (pA.east |- s1) -- (s1);
\draw[arr] (pA.east |- s2) -- (s2);
\draw[arr] (rc.east) -| (s1.north);
\draw[arr] (s1) -- (q1);
\draw[arr] (s2) -- (q2);
\draw[arr] (dB) -- (pB);
\draw[arr] (pB) -- (s3);
\draw[arr] (s3) -- (q3);
\draw[arr] (dC) -- (pC);
\draw[arr] (pC) -- (s4);
\draw[arr] (s4) -- (q4);
\end{tikzpicture}
}
\caption{Study Design. Pass A answers every scored question of the 2018--2025 exams and serves RQ1 and RQ2. Pass B asks for a reason per option on the questions with physician commentary (RQ3). Pass C repeats the image questions with the plate shown, withheld or swapped (RQ4).}
\label{fig:design}
\end{figure*}
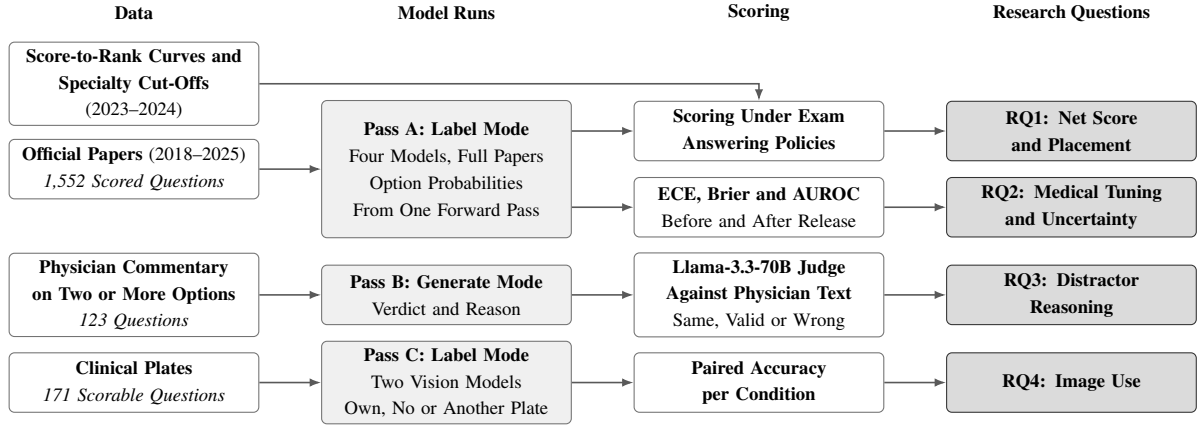

Figure~\ref{fig:design} summarizes the study design. Pass A answers every scored question of the 2018--2025 exams in label mode, and serves RQ1 and RQ2. Pass B asks for a reason per option on the questions with physician commentary (RQ3). Pass C repeats the image questions with the plate shown, withheld or swapped (RQ4). The four models are evaluated in parallel on each pass, and all runs are deterministic.

\subsection{Models}

We evaluate Qwen3-8B and Qwen3-14B \cite{Qwen3} together with Gemma-3-27B and MedGemma-27B \cite{Gemma3,MedGemma}, whose sizes and precisions are listed in Table~\ref{tab:models}. All four were released in 2025, before the 2025 exam was sat. Since MedGemma-27B is built on Gemma-3-27B through medical tuning, the pair isolates the effect of that tuning at a fixed scale and architecture, and as both accept images, they are also the two models evaluated in Pass C. All experiments ran on a single 48GB NVIDIA Quadro RTX 8000, the kind of hardware a hospital can realistically own and operate. Qwen3's thinking mode was disabled so that all four models answer directly.

\begin{table}[t]
\caption{Evaluated models and the judge.}
\label{tab:models}
\centering
\small
\begin{tabular}{@{}lrll@{}}
\toprule
Model & Params & Precision & Images \\
\midrule
Qwen3-8B & 8.2B & 16-bit & No \\
Qwen3-14B & 14.8B & 16-bit & No \\
Gemma-3-27B & 27.4B & 8-bit & Yes \\
MedGemma-27B & 27.4B & 8-bit & Yes \\
\midrule
Llama-3.3-70B (Judge) & 70.6B & 4-bit & No \\
\bottomrule
\end{tabular}
\end{table}

\subsection{Answering Protocol}

All prompts are written in Spanish, the language of the examination. The system prompt introduces the model as a MIR candidate and asks it to answer in the language of the question, while the user turn contains the stem and the numbered options. The model is not told about the scoring rule, since abstention is applied afterwards on the basis of its confidence (Section~\ref{sec:rq1-method}).

\medskip
\noindent\textbf{Label Mode.} In this mode, used for RQ1, RQ2 and RQ4, the model is asked only for the number of the correct option. In a single forward pass we read the probability $q_k$ that the model assigns to each option number $k$ as the first token of its reply and renormalize it over the options, $\hat{p}_k = q_k / \sum_j q_j$. The answer is then $\arg\max_k \hat{p}_k$ and the confidence $\max_k \hat{p}_k$, thus derived from the model's distribution over the options.

\medskip
\noindent\textbf{Generate Mode.} In this mode, used for RQ3, the model is asked for a reply stating, for every option, whether it is the one to mark together with a one sentence clinical reason, followed by the final answer and a stated confidence. Asking whether an option is \textit{``the one to mark''} rather than whether it is \textit{``true''} keeps the verdict unambiguous on negatively phrased stems (e.g., ``which of the following is FALSE''), which make up about a fifth of the MIR.

\subsection{Scoring and Placement (RQ1)}
\label{sec:rq1-method}

Each full-paper run is scored according to \eqref{eq:score} under three answering policies: \emph{always} hands in every answer, \emph{oracle} hands in only the correct ones and thus sets the ceiling for a perfectly self-aware model, and \emph{calibrated} applies the break-even rule of \eqref{eq:breakeven}. The latter requires an estimate of $p$ that the confidence cannot provide, because the largest of four probabilities summing to one is at least $1/4$, so the rule ``answer if $\max_k \hat{p}_k \geq 1/4$'' would never leave a blank. We therefore estimate $p$ from the model's record on its other seven examinations, sorting their answers by confidence into ten equal-sized groups and taking as $\hat{P}(\text{correct})$ the accuracy of the group into which a new answer's confidence falls, so that no examination's policy is fitted on its own answers. From these policies we derive the \emph{abstention gain}, calibrated minus always, and the \emph{headroom}, oracle minus calibrated, that is, the part of the net score that only a more accurate model could recover.

To place a model in the national ranking, we fit a \textit{score-to-rank} curve for each exam through the published points and interpolate linearly between the resulting knots. A specialty is considered \textit{open} at a given rank if its last place went to that rank or a later one, so the most competitive open specialty is the best choice still available, and the hospital that took its last place is where it remained available at the end. Mapping the 95\% interval of the net score through the curve then gives the corresponding range of ranks.

\subsection{Calibration (RQ2)}

For each model, we report accuracy, mean confidence, expected calibration error (ECE, with ten equal bins) \cite{Naeini2015}, the multi-class Brier score over the option distribution \cite{Brier1950}, and the area under the ROC curve (AUROC) with which confidence separates right from wrong answers \cite{Hendrycks2017}. Whereas ECE measures whether confidence matches accuracy on average, AUROC measures whether it ranks answers correctly. The main contrast is between MedGemma-27B and Gemma-3-27B on the same questions. As a contamination control, we also report every measure separately for the pre-release examinations, whose questions have circulated publicly for years, and for the 2025 exam, which was sat after all four models had been released.

\subsection{Rationale Grading (RQ3)}

Each model performs one \textit{generate mode} run over the 123 questions with commentaries on two or more options. For every option that has both a physician span and a model reason, an \textit{external judge}, i.e., a Llama-3.3-70B-Instruct \cite{Llama3} in 4-bit precision, is shown the question, the options, the key, the model's verdict on the option, the physician's text for it and the model's reason, and labels the reason as \emph{same} if it gives the physician's key clinical reason, \emph{valid} if it gives a different but clinically correct and sufficient one, or \emph{wrong} if it is clinically wrong, unsupported, irrelevant or too vague, or argues for the wrong verdict. Because the reasons are graded against the physician's commentary, RQ3 is limited to the 108 questions on which the physician's answer matches the exam key. Our main measure of interest, \emph{right answer, wrong reason}, is the share of correctly answered questions on which at least one distractor was rejected for a reason labelled wrong, with which we compare the \textit{label mode} confidence of these answers with that of correct answers whose distractor reasons were all sound.

\subsection{Image Conditions (RQ4)}

The two vision models answer the 171 image questions in \textit{label mode} under three conditions, i.e., with the question's own plate, with no plate, and with the plate of a different question. We report accuracy per condition, the paired difference between the real plate and each alternative, the share of answers that change, and the share of questions answered correctly in all three conditions, for which the text alone is evidently sufficient.

\subsection{Statistical Analysis}

All confidence intervals are 95\% intervals computed over questions, so they express how much a measure would change had a different set of questions been asked. Since every answer is deterministic, this is the only source of variation. For proportions, namely the accuracies and the RQ3 rate, we use Wilson intervals \cite{Wilson1927}. For the measures that are not simple proportions (e.g., net score, ECE, Brier score and AUROC), and for the difference between two values of any measure, we use a percentile bootstrap with 2,000 resamples of the questions \cite{Efron1993}. Two kinds of difference arise. When both values come from the same questions, as in MedGemma-27B minus Gemma-3-27B and in the real plate minus each alternative, the questions are resampled once and both values recomputed on them (a paired bootstrap). When they come from different questions, as in the 2025 exam minus the earlier ones and in the RQ3 confidence gap, the two sets are resampled independently.

As for significance tests, two models or two image conditions answering the same questions are compared with McNemar's exact test \cite{McNemar1947}, which considers only the questions on which the two disagree. The RQ3 rates, in contrast, are computed over each model's own correct answers and therefore over different questions, so we compare them with Fisher's exact test. Because this comparison involves all six pairs of the four models, its six $p$-values are corrected with Holm's adjustment \cite{Holm1979}. No other test in the paper is adjusted.

\section{\uppercase{Results}}
\label{sec:results}

\subsection{RQ1: Scored Like a Real Candidate}

\begin{figure*}[t]
\centering
\includegraphics[width=0.88\textwidth]{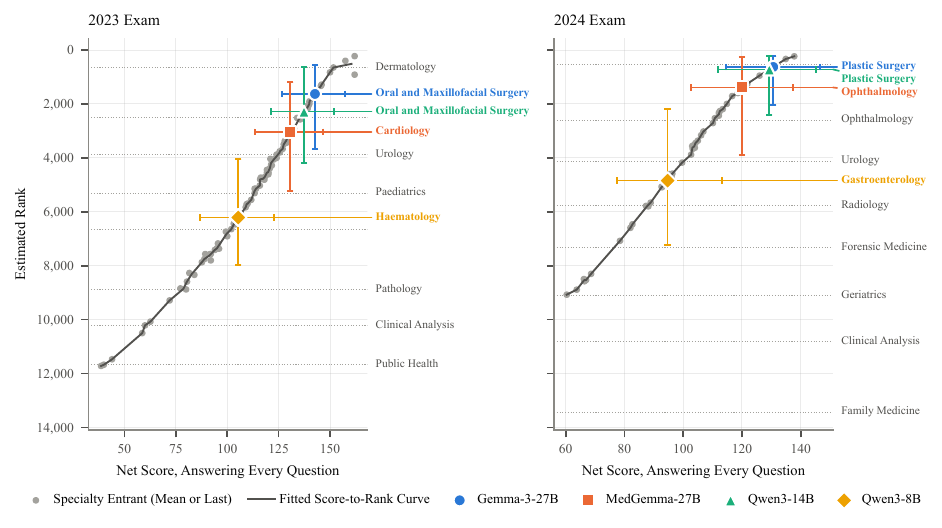}
\caption{RQ1: Each Model on the Score-to-Rank Curves of the 2023 and 2024 Exams, Answering Every Question. Grey points are the published per-specialty entrants; dotted lines mark a selection of specialty cut-offs, named in grey. Bars show the 95\% interval of the net score and the rank range it maps to. A solid line leads from each model to the right, where the most competitive specialty still open at its rank is named in the model's colour.}
\label{fig:match}
\end{figure*}

When answering every question, Gemma-3-27B and Qwen3-14B obtain an average net score per examination of 131.3 and 131.0, respectively, followed by MedGemma-27B with 125.5 and Qwen3-8B with 95.8, with a 95\% interval of roughly $\pm$16 on any single examination. Since the eight examinations differ in length, from 172 to 221 scored questions, Table~\ref{tab:rq1} reports the net score per 100 questions over all 1,552 of them, which is 67.7, 67.5, 64.7 and 49.4 for the same four models. Furthermore, Figure~\ref{fig:match} and Table~\ref{tab:place} show how these net scores map to the 2023 and 2024 \textit{score-to-rank} curves. In 2023, Gemma-3-27B's net score of 142.7 maps to an estimated rank of about 1,628, at which 43 of the 45 specialties were still open, the most competitive being Oral and Maxillofacial Surgery. Qwen3-14B lands close behind with the same choice, whereas MedGemma-27B would reach Cardiology and Qwen3-8B, with 22 specialties left, Haematology. In 2024 the three larger models reach ranks between about 600 and 1,400, with Plastic Surgery or Ophthalmology still open. These ranges are nonetheless wide, since a single examination places a model in the 95\% interval of roughly 2,000 ranks, and the 2024 curve, which is based on only 45 points, places any given score markedly higher than the 2023 curve (Section~\ref{sec:limitations}).

\begin{table}[t]
\caption{RQ1, pooled over the 2018--2025 exams (1,552 questions). Net score, gain and headroom per 100 questions. Gain is calibrated minus always, headroom oracle minus calibrated. Tenth: accuracy of the tenth of answers each model was least sure of.}
\label{tab:rq1}
\centering
\small
\setlength{\tabcolsep}{3.5pt}
\begin{tabular}{@{}lrrrrr@{}}
\toprule
Model & Acc. & Net & Gain & Headroom & Tenth \\
\midrule
Gemma-3-27B & .758 & 67.7 & 0.0 & 8.1 & .387 \\
MedGemma-27B & .735 & 64.7 & 0.0 & 8.8 & .368 \\
Qwen3-14B & .756 & 67.5 & 0.0 & 8.1 & .445 \\
Qwen3-8B & .621 & 49.4 & 0.0 & 12.7 & .303 \\
\bottomrule
\end{tabular}
\end{table}

\begin{table*}[t]
\caption{RQ1: Placement answering every question. Net score; estimated rank with the range that the 95\% interval of the net score maps to (233 is the best rank the 2024 curve can give); open specialties; and the most competitive specialty still open at that rank, with the hospital that took its last place (CHU: Complejo Hospitalario Universitario; CH: Complejo Hospitalario; CAU: Complejo Asistencial Universitario; H. U.: Hospital Universitari).}
\label{tab:place}
\centering
\footnotesize
\setlength{\tabcolsep}{3pt}
\begin{tabular}{@{}lrrrll@{}}
\toprule
Model & Score & Rank & Open & Best Open Specialty & Its Last Place \\
\midrule
\multicolumn{6}{@{}l}{\textbf{2023 Exam (196 questions)}} \\
Gemma-3-27B & 142.7 & 1,628 (561--3,687) & 43/45 & Oral and Maxillofacial Surgery & CHU de A Coruña \\
Qwen3-14B & 137.3 & 2,284 (643--4,189) & 43/45 & Oral and Maxillofacial Surgery & CHU de A Coruña \\
MedGemma-27B & 130.7 & 3,029 (1,198--5,226) & 41/45 & Cardiology & Hospital de Mérida \\
Qwen3-8B & 105.3 & 6,212 (4,058--7,962) & 22/45 & Haematology & H. U. Arnau de Vilanova, Lleida \\
\midrule
\multicolumn{6}{@{}l}{\textbf{2024 Exam (200 questions)}} \\
Gemma-3-27B & 130.7 & 626 (233--2,025) & 43/44 & Plastic Surgery & CHU de Cáceres \\
Qwen3-14B & 129.3 & 719 (233--2,398) & 43/44 & Plastic Surgery & CHU de Cáceres \\
MedGemma-27B & 120.0 & 1,387 (250--3,883) & 42/44 & Ophthalmology & CH de Mérida \\
Qwen3-8B & 94.7 & 4,843 (2,201--7,219) & 30/44 & Gastroenterology & CAU de Palencia \\
\bottomrule
\end{tabular}
\end{table*}

\begin{figure}[t]
\centering
\includegraphics[width=\columnwidth]{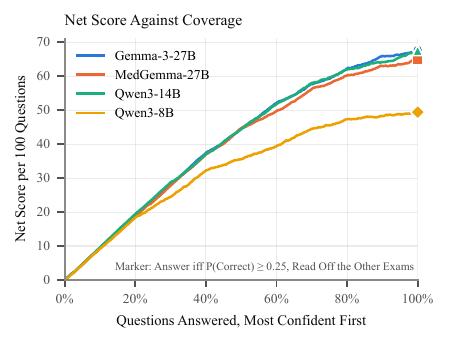}
\caption{RQ1: Net Score per 100 Questions Against the Share Answered, Most Confident First, Pooled Over 2018--2025.}
\label{fig:selective}
\end{figure}

Abstention, on the other hand, never helps. The calibrated policy answers every question on every examination, so the abstention gain is zero for all 32 model-examination pairs. This is not because confidence is uninformative, since it separates right from wrong answers with an AUROC of 0.80 for the three larger models and 0.76 for Qwen3-8B (Table~\ref{tab:rq2}), but because even the least confident tenth of each model's answers on its other examinations was right 29--47\% of the time, above the 25\% threshold of \eqref{eq:breakeven}. Indeed, skipping the least confident tenth chosen in hindsight within each examination would have lowered the net score on 28 of the 32 pairs and gained at most 3.7 on the others. Figure~\ref{fig:selective} illustrates why: the net score rises almost monotonically as answers are added from most to least confident. The net score lost to wrong answers, 15.7--24.5 per examination (8.1--12.7 per 100 questions), is thus \textit{headroom} that a more accurate model could recover.

\subsection{RQ2: Medical Tuning and Uncertainty}

\begin{figure}[t]
\centering
\includegraphics[width=0.9\columnwidth]{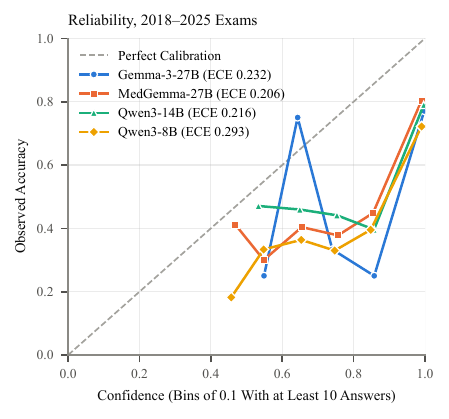}
\caption{RQ2: Reliability Diagram, Pooled Over the 2018--2025 Exams. Only bins with at least ten answers are drawn; most answers fall in the top bin.}
\label{fig:reliability}
\end{figure}

All four models are overconfident (Figure~\ref{fig:reliability}), with a mean confidence of 0.91--0.99 against an accuracy of 0.62--0.76 and an ECE of 0.21--0.29 (Table~\ref{tab:rq2}). Below a confidence of 0.9, in fact, accuracy falls under 0.5 in all but one of the bins with at least ten answers. Medical tuning, however, changed the level of confidence. Over all eight examinations, MedGemma-27B is 2.3 points less accurate than Gemma-3-27B on the same questions ($p = 0.027$) and 0.047 less confident, which lowers its ECE by 0.026 [0.005, 0.045]. Its AUROC, however, is essentially unchanged, with a difference of +0.002 [$-$0.025, +0.028], so it is no better at telling its right answers from its wrong ones. Given that no confidence-based abstention pays on this examination (RQ1), the lower ECE brings no gain in net score, whereas the loss of accuracy costs about 3 points of net score per 100 questions.

\begin{table*}[t]
\caption{RQ2: Calibration before release (Pre: the 2018--2024 exams, 1,358 questions), after release (2025, 194 questions) and over both (All), with the paired difference MedGemma-27B minus Gemma-3-27B. $p$ is McNemar's exact test on accuracy.}
\label{tab:rq2}
\centering
\footnotesize
\setlength{\tabcolsep}{3pt}
\begin{tabular}{@{}llrrrrrr@{}}
\toprule
Model & Exams & Accuracy & Confidence & ECE & Brier & AUROC & $p$ \\
\midrule
Gemma-3-27B & Pre & .756 & .988 & .234 [.212, .257] & .472 & .792 [.766, .818] & \\
 & 2025 & .768 & .990 & .222 [.166, .283] & .448 & .854 [.790, .910] & \\
MedGemma-27B & Pre & .729 & .941 & .214 [.194, .237] & .469 & .799 [.776, .823] & \\
 & 2025 & .778 & .941 & .163 [.117, .219] & .359 & .812 [.731, .883] & \\
Qwen3-14B & Pre & .752 & .972 & .221 [.200, .244] & .464 & .790 [.765, .815] & \\
 & 2025 & .789 & .975 & .186 [.135, .240] & .369 & .866 [.803, .921] & \\
Qwen3-8B & Pre & .616 & .913 & .297 [.274, .322] & .650 & .748 [.723, .775] & \\
 & 2025 & .655 & .915 & .261 [.204, .322] & .571 & .830 [.770, .884] & \\
\midrule
Difference & Pre & $-$.027 [$-$.049, $-$.006] & $-$.046 & $-$.020 [$-$.041, +.002] & $-$.003 & +.006 [$-$.024, +.036] & .013 \\
 & 2025 & +.010 [$-$.041, +.057] & $-$.049 & $-$.059 [$-$.105, $-$.011] & $-$.089 & $-$.042 [$-$.128, +.038] & .84 \\
 & All & $-$.023 [$-$.043, $-$.004] & $-$.047 & $-$.026 [$-$.045, $-$.005] & $-$.014 & +.002 [$-$.025, +.028] & .027 \\
\bottomrule
\end{tabular}
\end{table*}

\begin{figure*}[t]
\centering
\includegraphics[width=0.9\textwidth]{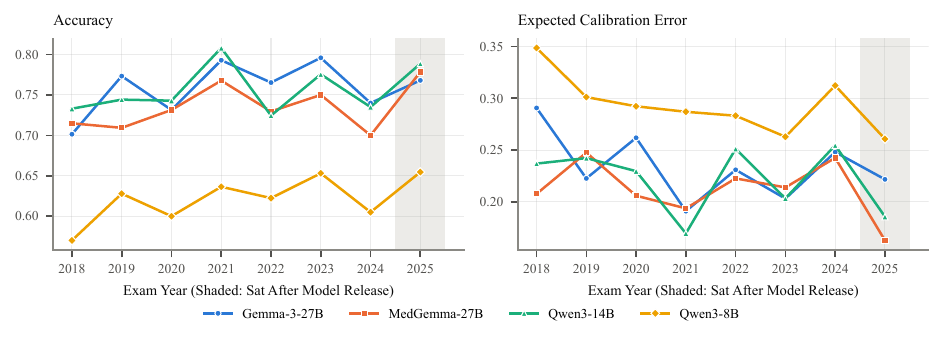}
\caption{RQ2: Accuracy and Expected Calibration Error by Exam Year. The 2025 exam was sat after all models released.}
\label{fig:contamination}
\end{figure*}

Moreover, we find no sign of \textit{memorization} (Figure~\ref{fig:contamination}). On the 2025 exam, which none of the models may have seen during training, all are at least as accurate as on the older exams, by 1.2 to 4.9 points (Table~\ref{tab:rq2}), and its mean confidence matches that on the older exams to within 0.003. If anything, the Qwen models rank their answers better on 2025, with AUROC differences of +0.075 [+0.007, +0.138] and +0.082 [+0.010, +0.143]. Likewise, the accuracy gap between Gemma and MedGemma is no smaller on 2024 (4.0 points) than on 2018--2023 (2.5 points). On the 2025 exam the two models are equally accurate while MedGemma's ECE advantage grows to 0.059, although a single post-release exam cannot distinguish an easier one from a less memorized one.

\subsection{RQ3: Right Answers, Wrong Reasons}

\begin{figure*}[t]
\centering
\includegraphics[width=0.9\textwidth]{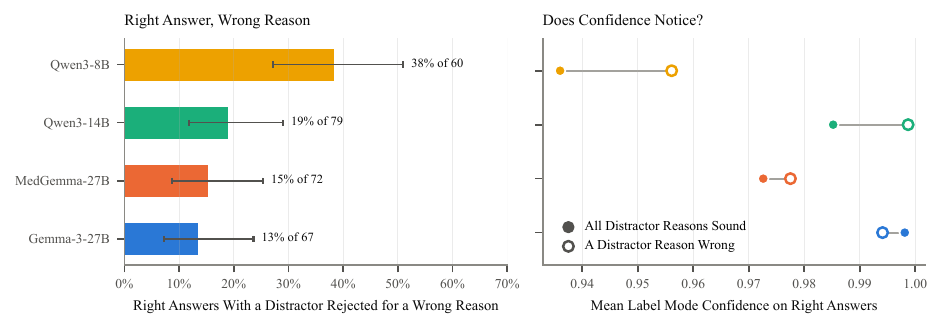}
\caption{RQ3: Share of Correct Answers With at Least One Distractor Rejected for a Wrong Reason, With 95\% Wilson Intervals (Left), and Mean Label Mode Confidence on Those Answers Against Correct Answers Whose Distractor Reasons Were All Sound (Right).}
\label{fig:rationale}
\end{figure*}

A correct answer frequently rests on a \textit{distractor} rejected for a wrong reason (Figure~\ref{fig:rationale}, Table~\ref{tab:rq3}). This occurs on 38\% [27, 51] of the correct answers of Qwen3-8B, compared with 19\% [12, 29] for Qwen3-14B, 15\% [9, 25] for MedGemma-27B and 13\% [7, 24] for Gemma-3-27B. After Holm's adjustment, Qwen3-8B differs from Gemma-3-27B ($p = 0.011$) and from MedGemma-27B ($p = 0.024$) and falls just short of significance against Qwen3-14B (unadjusted $p = 0.013$, adjusted 0.053), while no other pair differs. In particular, Gemma-3-27B and MedGemma-27B do not ($p = 0.81$ before adjustment), so medical tuning did not change how often distractors are dismissed for wrong reasons. Across all distractor reasons, the judge labels 64--74\% as valid, 14--18\% as the same as the physician's and 11--21\% as wrong.

\begin{table}[t]
\caption{RQ3 on 108 questions. Acc.: generate mode accuracy, with label mode accuracy on the same questions in brackets. Right: correct answers with graded distractors. Gap: label mode confidence on \textit{wrong reason} minus \textit{sound reason} correct answers.}
\label{tab:rq3}
\centering
\small
\setlength{\tabcolsep}{2.5pt}
\begin{tabular}{@{}lrrrr@{}}
\toprule
Model & Acc. & Right & Wrong Reason & Gap \\
\midrule
Qwen3-8B & .56 (.60) & 60 & .38 & +.020 \\
Qwen3-14B & .73 (.77) & 79 & .19 & +.014 \\
MedGemma-27B & .67 (.71) & 72 & .15 & +.005 \\
Gemma-3-27B & .62 (.77) & 67 & .13 & $-$.004 \\
\bottomrule
\end{tabular}
\end{table}

Moreover, the models' confidence does not reflect these errors. The confidence of correct answers reached with a wrongly rejected distractor is no lower than that of correct answers with sound reasons, the difference ranging from $-$0.004 to +0.020 (Table~\ref{tab:rq3}) with only Qwen3-14B's interval, [+0.002, +0.031], excluding zero, and in the wrong direction. The stated confidence in generate mode does not separate them either. Likewise, asking for a reason per option also changes the answers themselves. On the same 108 questions, generate mode is 4--5 points less accurate than label mode for three of the models and 15 points less accurate for Gemma-3-27B (from 0.77 to 0.62).

Table~\ref{tab:examples} illustrates the judge's grading of distractor reasons. In the first two examples the model's reason is \textit{clinically wrong}, since pulmonary fibrosis is a classic cause of digital clubbing and diffuse idiopathic skeletal hyperostosis is diagnosed radiographically. The third example, in contrast, shows a judge error: on a \textit{``which statement is true''} stem, the model rejects the statement that proteinuria above 500mg/24h is unlikely by arguing that proteinuria is in fact likely, which is also the physician's reason, and yet the judge labels this wrong for three of the four models. The judge can thus misread a reason that negates a negatively worded option, which inflates the wrong-reason rate for every model.

\begin{table*}[t]
\caption{Graded distractor reasons on questions the model answered correctly, translated from Spanish and abridged.}
\label{tab:examples}
\centering
\scriptsize
\setlength{\tabcolsep}{4pt}
\begin{tabular}{@{}>{\raggedright\arraybackslash}p{4.5cm}>{\raggedright\arraybackslash}p{2.3cm}>{\raggedright\arraybackslash}p{3.1cm}>{\raggedright\arraybackslash}p{3.6cm}>{\raggedright\arraybackslash}p{1cm}@{}}
\toprule
Question and Key & Option Rejected & Physician & Model Reason & Judge \\
\midrule
Ex-smoker with COPD and new digital clubbing; most plausible cause? Key: Lung carcinoma (MIR 2015, No.~125). & Pulmonary fibrosis. & The other conditions can also cause it, but not so suddenly. & MedGemma-27B: ``Pulmonary fibrosis is not usually associated with digital clubbing.'' & Wrong, correctly. \\
\addlinespace
70 y.o., a week of progressive low back pain and low-grade fever, normal radiograph; what to rule out first, and how? Key: Pyogenic spondylodiscitis, MRI (2019, No.~122). & Ankylosing vertebral hyperostosis, with a dorsolumbar radiograph. & Vertebral collapse, ankylosing hyperostosis and many metastases would be seen on the plain radiograph. & Qwen3-8B: ``Radiography is insufficient to diagnose ankylosing hyperostosis, which requires more specific studies.'' & Wrong, correctly. \\
\addlinespace
Diabetic nephropathy, eGFR 14~ml/min; which statement is true? Key: Stage 5, consider renal replacement (2021, No.~149). & Proteinuria above 500~mg/24~h is unlikely. & These patients usually have A3 albuminuria, possibly nephrotic-range (option false). & Gemma-3-27B: ``Proteinuria is common in diabetic nephropathy and likely above 500~mg/24~h at such an advanced stage.'' & Wrong, judge error. \\
\bottomrule
\end{tabular}
\end{table*}

\subsection{RQ4: Use of the Clinical Image}

\begin{figure}[t]
\centering
\includegraphics[width=\columnwidth]{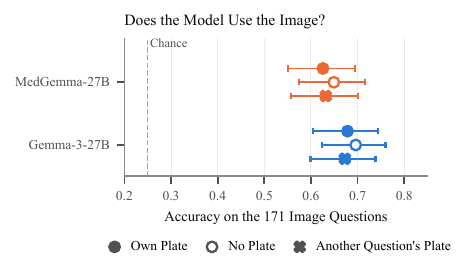}
\caption{RQ4: Accuracy of the Vision Models on the 171 Image Questions With the Question's Own Plate, No Plate and Another Question's Plate, With 95\% Wilson Intervals.}
\label{fig:images}
\end{figure}

The clinical images available along questions make no measurable difference in terms of model accuracy (Figure~\ref{fig:images}). Gemma-3-27B answers 67.8\% of the 171 image questions correctly with the real plate, 69.6\% with no plate and 67.3\% with another question's plate, and MedGemma-27B 62.6\%, 64.9\% and 63.2\%, respectively. Relative to no plate, the real plate thus changes accuracy by $-$1.8 points [$-$7.0, +2.9] and $-$2.3 points [$-$7.6, +2.4] ($p = 0.65$ and $0.50$), so it adds at most about 3 points, and it stays within 0.6 points of a mismatched plate ($p = 1.0$ for both). The plate is not ignored outright, since it changes 13--20\% of the answers, but the gains and losses cancel each other out, and 58\% (Gemma) and 56\% (MedGemma) of the image questions are answered correctly in all three conditions, i.e., from the text alone.

\subsection{Accuracy by Specialty}

\begin{figure*}[t]
\centering
\ifcameraready
\includegraphics[width=0.95\textwidth]{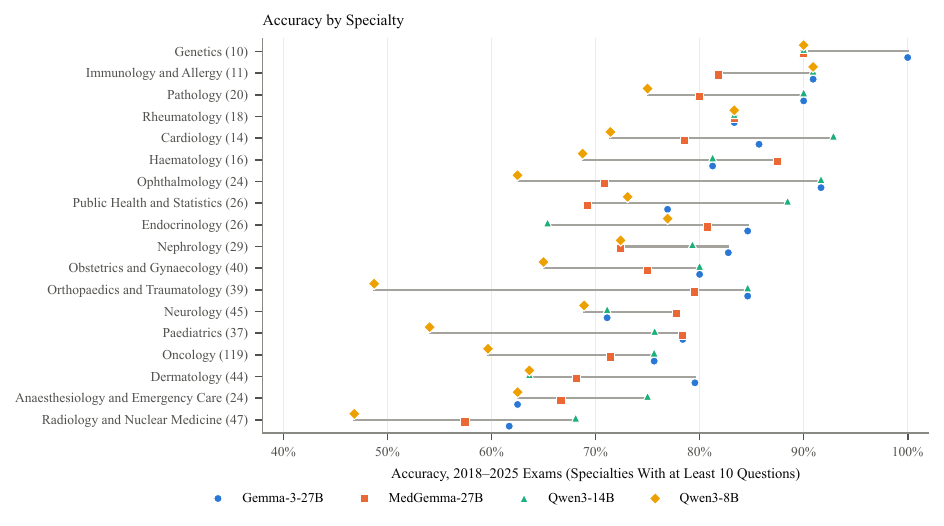}
\else
\includegraphics[width=0.68\textwidth]{figures/fig_specialty.pdf}
\fi
\caption{Accuracy by Specialty Over the 2018--2025 Exams, for the 18 Specialties With at Least Ten Labelled Questions (Question Count in Brackets). The grey line spans the four models and markers are offset so that equal accuracies stay visible.}
\label{fig:specialty}
\end{figure*}

Figure~\ref{fig:specialty} breaks accuracy down for the 18 specialties with at least ten labelled questions, which together account for 589 questions. Radiology and Nuclear Medicine is the hardest specialty for every model except Qwen3-14B, with accuracies of 47--68\%, whereas the smallest model falls furthest behind in Orthopaedics and Traumatology (49\% against 80--85\%) and in Paediatrics (54\% against 76--78\%). Medical tuning does not produce a distinct specialty profile either, as MedGemma-27B is less accurate than Gemma-3-27B in 13 of the 18 specialties, ties in two, and is ahead only in Haematology, Neurology and Anaesthesiology, by at most 7 points. With 10 to 119 questions per specialty, however, single differences amount to a few questions only.

\section{\uppercase{Discussion}}
\label{sec:discussion}

\noindent\textbf{Penalty Is Too Mild to Reward Self-Knowledge.} Penalized scoring has been proposed as a remedy for benchmarks that reward guessing \cite{Kalai2025}, and the MIR illustrates the limits of this approach. With a reward $a$ for a right answer and a penalty $b$ for a wrong one, answering every question is optimal whenever $p > b/(a+b)$, i.e., 0.25 for the MIR. Models whose least confident answers are right a third of the time therefore never benefit from abstaining, however good their confidence may be, and the examination cannot distinguish a model that knows what it does not know from one that does not. A clinical deployment, in contrast, faces much steeper penalties, since a wrong recommendation that is acted upon can cost far more than a deferral. In that setting the break-even rises towards one, and the ability of confidence to rank errors, here an AUROC of about 0.80, becomes decisive and should be reported alongside.

\medskip
\noindent\textbf{Medical Tuning Is Not a Safety Proxy.} In the absence of a penalty that rewards self-knowledge, a hospital may be tempted to assume that a medical model is safer than its base model. In our experiments, MedGemma-27B was less overconfident than its base model, which may be taken as a sign of better self-knowledge. However, the improvement came from lowering confidence across the board rather than from a better knowledge of which answers were wrong, and it came at a cost in \textit{accuracy}. A hospital choosing between a general model and its medical variant should thus test both on \textit{task} and on \textit{language}, rather than assume that the medical label brings better judgement.

\medskip
\noindent\textbf{Explanations Need to Be Checked.} When a model's explanation is shown to a clinician or a student, between 13 and 38\% of its correct answers come with a distractor dismissed for a wrong reason. In medical education, where the explanation is the main object of interest, this is a serious flaw, and in clinical deployment, it is a safety concern. The model's confidence does not reveal it, and the judge's grading shows that the problem is not limited to a single model. Asking for explanations also changed the answers, by 15 points for one model, so an evaluation in which the model is asked to explain its answer may not reflect its performance when it is not.

\medskip
\noindent\textbf{Images Are Optional on This Examination.} More than half of the image questions were answered correctly without any image, and neither model used the images to improve accuracy. For clinical use, this implies that the ability to accept an image is no evidence that the model ultimately \textit{looks} at it, and that image dependent tasks require their own evaluation.

\section{\uppercase{Limitations}}
\label{sec:limitations}

\noindent\textbf{Judge Validation.} No physician has rated the judge's grading, and Table~\ref{tab:examples} shows that the judge errs on reasons that negate negatively worded options. Physician ratings of a blind sample, as well as the expert ratings of LLM reasoning available for rheumatology questions \cite{MadridGarcia2023}, would provide a more reliable estimate. In addition, RQ3 excludes the 15 questions whose physician answer differs from the exam key, and covers pre-release examinations only.

\medskip
\noindent\textbf{Ranks Are Estimates.} Since the rank combines the examination with the academic record, a rank read off the net score assumes a \textit{typical} record. Moreover, the 2024 curve rests on specialty means only and, at an equal score, yields a much better rank than the 2023 curve, so its placements should be read as \textit{optimistic}.

\medskip
\noindent\textbf{Scope.} Our study covers four open models of 8--27B parameters, a single prompt in Spanish and direct answering, so larger or reasoning models may behave differently. The Gemma pair ran in 8-bit precision and the Qwen models in 16-bit, although the Gemma models were tested in equal conditions. Confidence is measured as a probability over the options, and verbalized or sampled confidence may behave differently. Finally, images are available only from 2019 onwards, and the 2017 exam is excluded from RQ1, RQ2 and RQ4 because of parsing issues.

\section{\uppercase{Conclusions}}
\label{sec:conclusions}

Scored as real candidates, open models that a hospital can run on its own hardware would pass the Spanish residency examination with room to spare and could choose almost any specialty. That score, however, reveals little of what a health system needs to know before relying on them. The exam's penalty is too mild for any of these models to benefit from abstaining, so it cannot reward a model for knowing its limits, and medical tuning lowered confidence without improving the models' self-knowledge, and it cost accuracy. Likewise, up to 38\% of correct answers rested on distractors rejected for a wrong reason, which the models' confidence did not reveal, and the provided clinical images made no difference to accuracy. Taken together, these show that a model can match into a residency while being overconfident, right for the wrong reasons, and indifferent to the images it is shown. Evaluations intended to support clinical deployment should therefore look beyond the score, reporting how well confidence ranks errors, verifying the reasoning behind correct answers, and testing the use of images directly. Validating the rationale judge against physician ratings and extending the analysis to larger and reasoning models are natural next steps.

\ifcameraready
\section*{\uppercase{Acknowledgements}}

This work was supported by the Basque Government under grant DEUSTEK6 -- Humanized Computing for Smart Sustainable and Healthier Communities and Environments (IT1901-26), and by the European Union's Horizon Europe Research and Innovation Programme under the LATE-AYA project (Grant Agreement No.~101214326).
\fi

\bibliographystyle{apalike}
{\small
\bibliography{mir_reasoning}}

\end{document}